\documentclass{Style/ARXIV}

\interspeechcameraready

\usepackage{xspace}
\usepackage{multirow}

\newcommand{\goal}{\mathcal{G}}
\newcommand{\instruction}[1]{\mathcal{I}_{#1}}
\newcommand{\werr}[1][\relax]{\ifx\relax#1{$\Delta_{WER}$}\else{Relative Change in WER}\fi\xspace}
\newcommand{\rri}[1][\relax]{\ifx\relax#1{$\Delta_{RR}$}\else{Relative Change in RR}\fi\xspace}

\title{Multimodal Speech Recognition for Language-Guided Embodied Agents}
\begin{document}

\name{
Allen Chang$^{1,2}$, 
Xiaoyuan Zhu$^{1,2}$,
Aarav Monga$^{1,2}$,
Seoho Ahn$^{1,2}$,\\
Tejas Srinivasan$^{1}$, 
Jesse Thomason$^1$
}
\newcommand{\eqcontrfootnote}{
\let\thefootnote\relax\footnotetext{*Equal contribution}
}
\address{
  $^1$Department of Computer Science, University of Southern California, USA\\
  $^2$Center for Artificial Intelligence in Society, University of Southern California, USA}
\email{
changall
@usc.edu
}
\maketitle

\newcommand{\ic}[1]{
\ifdefined\DEBUG
\ifdefined\ANNOTATIONS
    \begin{mdframed}
    [linecolor=RubineRed,
     linewidth=2pt,
     topline=false,
     rightline=false,
     bottomline=false]
    \color{RubineRed}
    #1
    \end{mdframed}

\fi
\fi
}
\DeclareRobustCommand{\ann}[1]{\ifdefined\DEBUG{\sethlcolor{RubineRed}\hl{#1}}\else#1\fi}
\DeclareRobustCommand{\todo}[1]{\ifdefined\DEBUG{\hl{#1}}\else#1\fi}
\DeclareRobustCommand{\fc}[1]{\ifdefined\DEBUG{\sethlcolor{BurntOrange}\hl{#1}}\else#1\fi}
\DeclareRobustCommand{\cn}{\ifdefined\DEBUG{$^{[\text{\hl{CITE}}]}$}\fi}


\begin{abstract}
Benchmarks for language-guided embodied agents typically assume text-based instructions, but deployed agents will encounter spoken instructions. 
While Automatic Speech Recognition (ASR) models can bridge the input gap, erroneous ASR transcripts can hurt the agents’ ability to complete tasks. 
We propose training a multimodal ASR model that utilizes the accompanying visual context to reduce errors in spoken instruction transcripts. 
We train our model on a dataset of synthetic spoken instructions, derived from the ALFRED household task dataset, where we simulate acoustic noise by systematically masking spoken words. 
We find that utilizing visual observations facilitates masked word recovery, with multimodal ASR models recovering up to 30\% more masked words than unimodal baselines. 
We also find that spoken instructions transcribed by multimodal ASR models result in higher task completion success rates for a language-guided embodied agent.
\href{https://github.com/Cylumn/embodied-multimodal-asr}{\texttt{github.com/Cylumn/embodied-multimodal-asr}}
\end{abstract}
\noindent\textbf{Index Terms}: speech recognition, multimodal learning, human-robot interaction, embodied learning
\section{Introduction}

Several benchmarks aim to train simulation agents to complete household chores~\cite{shridhar2020alfred, puig2018virtualhome, padmakumar2022teach}.
These agents are trained to receive and follow written instructions, neglecting the more realistic scenario of receiving spoken instructions. 
However, physical hardware agents and robots require the ability to understand spoken language for effective interaction with human users~\cite{marge2022spoken}. 

\begin{figure}
    \centering
    \includegraphics[width=0.9\linewidth]{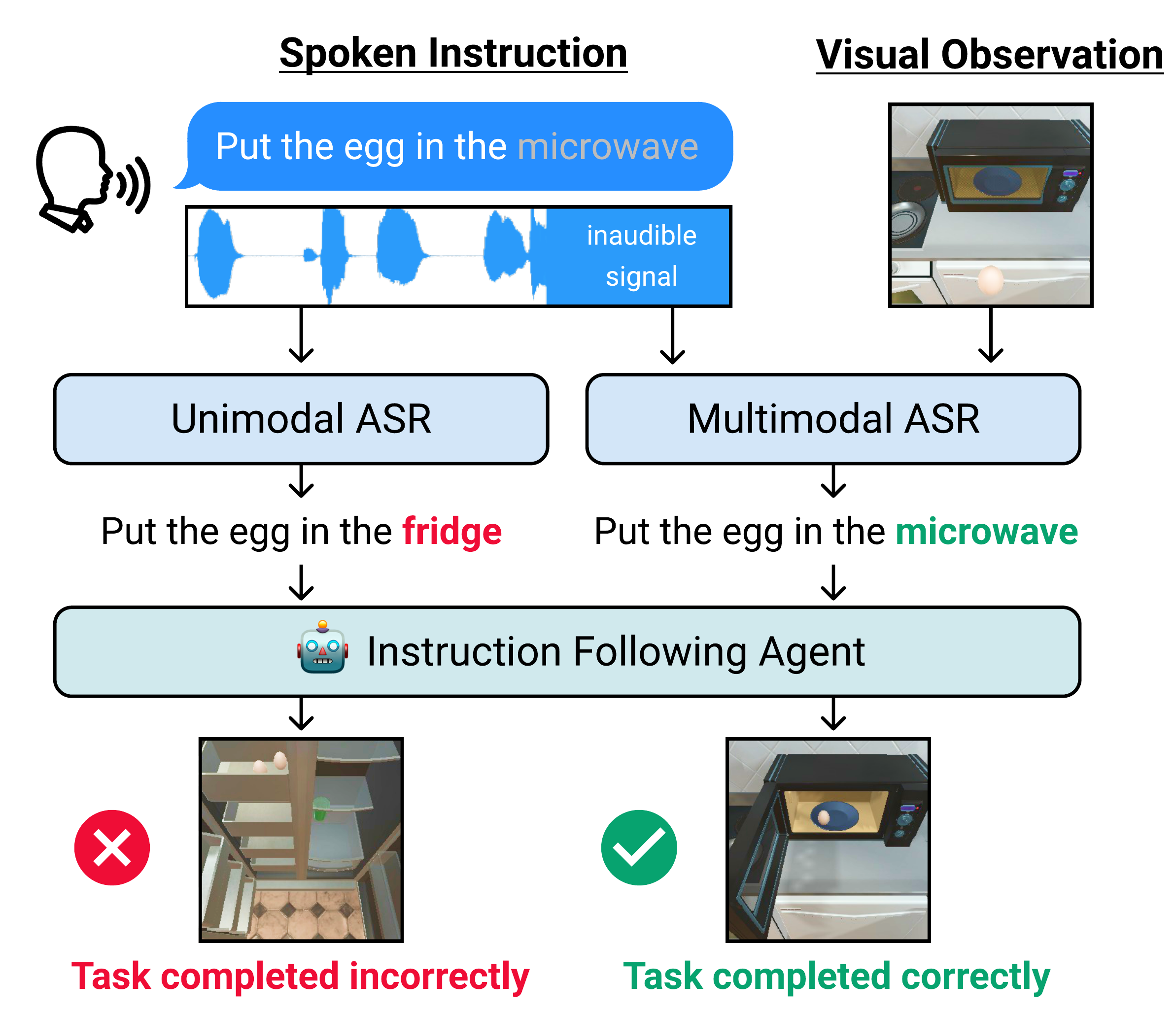}
    \caption{Multimodal ASR systems can leverage an embodied agent's visual observation to better transcribe spoken instructions instead of relying on language priors alone.}
    \label{fig:masr_example}
\end{figure}

Automatic Speech Recognition (ASR) methods can transcribe spoken instructions into written text. 
However, erroneous ASR transcripts can hurt the embodied agents' ability to complete the required tasks. 
One proposed solution for improving ASR robustness is to use visual context, when available, to recover inaudible words. 
Previous work studied situations where spoken language described a single image~\cite{srinivasan2020looking}.
We hypothesize that the visual observations made by embodied agents during task completion will be similarly helpful for disambiguating inaudible words.
In particular, we test whether visual observations from an embodied agent can be utilized by multimodal ASR models to reason about and recover words that are masked in the audio signal. 
Consider transcribing \textit{``Put the egg in the \dots''} where the destination word is too noisy to understand (Figure~\ref{fig:masr_example}).
A unimodal ASR model must use language priors alone to guess, for example, \textit{``fridge.''}
By contrast, a multimodal ASR model can use the agent's visual observation to correctly reason that the user wants the egg to be put in the \textit{``microwave.''}

We apply the insight that visual scene information can inform noisy spoken language transcription in the context of an embodied, instruction-following agent.
We synthesize a benchmark of text-to-speech (TTS) generated spoken instructions based on written instructions in the Action Learning From Realistic Environments and Directives (ALFRED) dataset~\cite{shridhar2020alfred}. 
To simulate acoustic noise, we systematically mask audio segments corresponding to words in the instructions.
We train unimodal and multimodal ASR models on spoken instructions and evaluate their ability to recover masked words in seen versus unseen visual environments and heard versus unheard TTS speakers.
We quantify the downstream impact of erroneous ASR transcriptions on an off-the-shelf ALFRED embodied agent.

We demonstrate that multimodal models can effectively use visual observations to transcribe noisy spoken instructions. 
We find that the additional visual context is useful in both seen and unseen household environments. 
Visual context also can help the ASR model generalize better to new speakers. 
We demonstrate that multimodal ASR models can mitigate the effect of noise in spoken instructions on a text-trained embodied agent's ability to complete tasks. 
These findings are promising for building embodied agents that follow spoken instructions.

\section{Background}
This work addresses a gap in the embodied task completion literature.
Most embodied agent training assumes clean text input, but human speech will be encountered in deployed settings.
We leverage insights from existing ASR literature to develop a multimodal ASR model for noisy spoken instructions.

\textbf{Embodied Task Completion.}
Existing benchmarks aim to train embodied agents to complete tasks by following instructions~\cite{gu2022vision}.
These benchmarks range from vision-language navigation~\cite{anderson2018vision,qi2020reverie,chen2019,thomason2020vision,banerjee:corl20}, where an agent must follow the user's instructions by executing actions to reach the desired destination, to embodied task completion~\cite{shridhar2020alfred,puig2018virtualhome,suhr:emnlp19,padmakumar2022teach}, where agents interact with objects in the environment to complete the user's instructions.
These benchmarks provide language as written text, and while a small number of benchmarks involve spoken instructions~\cite{rxr}, almost all modeling attempts assume text-based instructions.
There are some works at the intersection of embodied learning and acoustic signals~\cite{chen2020soundspaces,paulavlen}, but these do not involve spoken instructions. 
We explore multimodal ASR modeling for language-guided embodied agents.

\textbf{Multimodal Speech Recognition.}
Previous works have explored augmenting speech recognition with visual information such as lip readings~\cite{319179,duchnowski94_icslp}, visual scenes~\cite{miao2016open,palaskar2018end,sanabriahow2}, and task semantics~\cite{corona:ijcnlp17}.
When the audio signal is clear, the visual modality has been shown to regularize the model~\cite{caglayan2019multimodal} rather than assisting with semantic disambiguation.
In contrast, visual semantics have been shown to be helpful when the audio signal is degraded~\cite{ srinivasan2020looking,srinivasan2019analyzing}. 
When visually salient words are masked in the audio signal, multimodal ASR can utilize the visual input to recover the masked words~\cite{srinivasan2020fine}. 
We apply these findings to an embodied setting, where agents receiving instructions with noisy or degraded speech can leverage their visual observations to recover the instruction text and complete the requested task.

\section{Methodology}
We adapt the ALFRED~\cite{shridhar2020alfred} language-guided instruction benchmark to create synthesized speech commands, apply noising policies to that speech, and attempt to recover ground truth instructions using a novel, multimodal ASR model.

\subsection{Preliminaries: ALFRED Instruction Following Task}

ALFRED~\cite{shridhar2020alfred} is a benchmark in which an embodied agent must follow language instructions to complete tasks by navigating a room and interacting with objects.
Tasks consist of a language goal $\goal$ (e.g., \textit{``Find an egg in the fridge and microwave it.''}), a sequence of $K$ sub-goal instructions $\instruction{1...K}$ to achieve that goal, and target environment state conditions (e.g., an egg has been heated).
Training and validation tasks include an annotated sequence of actions to accomplish the goal.
At each timestep, the agent receives a single-frame visual observation $v \in \mathbb{R}^{W \times H \times 3}$ and executes a discrete navigation step (e.g., \texttt{RotateRight}) or object interaction (e.g., \texttt{PickUp(Egg)}) which updates the environment state.
The task is successfully completed when the environment state satisfies the target state conditions.

\subsection{Building a Dataset of Spoken Instructions}
We create a dataset of synthetic spoken instructions because of the lack of embodied learning datasets with speech inputs.

We extract sub-goal text instructions $\instruction{}$ from the ALFRED dataset and apply off-the-shelf TTS models\footnote{TTS models are sourced from the open-source Silero~\cite{silero} library.} to generate synthetic speech instructions $S(\instruction{})$.
We pair each instruction with a visual context $v$, the agent's observation when the previous sub-goal is completed.
For example, if the agent has navigated to a microwave as the previous subgoal, the observation accompanying \textit{``Put the egg in the microwave''} will be of a kitchen counter with a microwave (Figure~\ref{fig:masr_example}).
Multimodal ASR models trained on our data are tasked with transcribing the speech input $S(\instruction{})$ into the word sequence $\instruction{}$ using the paired visual context.

Since the test examples in the ALFRED dataset are not paired with annotated visual observations, we use the ALFRED validation data as our test data.
Our test set is partitioned into environments that were \textit{seen} and \textit{unseen} during training, enabling us to study the effects of in- versus out-of-distribution visual contexts on multimodal ASR.
We partition the ALFRED training data in a 90:10 split to generate our train and validation sets. 
Training instructions with exact text matches in our validation or test sets are removed.
In total, there are $41,474$ spoken instructions in the training set, $5,222$ in validation, $5,570$ in \textit{seen} test environments, and $5,140$ in \textit{unseen} test environments.

To examine the effects of multimodal ASR when generalizing across speakers, we create two sets of spoken instructions:
(1) we generate spoken instructions for training, validation, and test folds using a single American English speaker TTS model, denoted as $S_A(\instruction{})$; (2) we use 10 Indic English speaker TTS models for training and validation, the same 10 speakers for a \textit{heard} ASR test instruction split,  $S_I^{\text{heard}}(\instruction{})$, and 5 different Indic English speakers for an \textit{unheard} ASR test instruction split, $S_I^{\text{unheard}}(\instruction{})$.
In all cases, TTS outputs for all sub-goal instructions within a single task are generated from the same speaker.

\subsection{Injecting Noise into Spoken Instructions}

Prior studies in machine translation~\cite{caglayan2019probing} and ASR~\cite{srinivasan2020looking} have shown that visual context is helpful when the primary modality is degraded. 
Building on these insights, we corrupt the audio in our dataset by masking audio segments in the instructions. 

Following prior work~\cite{srinivasan2020looking}, we apply masking at the word-level.
After segmenting instructions with wav2vec 2.0~\cite{baevski2020wav2vec} to identify word boundaries, we mask words by substituting their speech segments with Gaussian noise. 
In our controlled setting, we can mask out different sets of words to evaluate the utility of the visual context under different conditions. 

\begin{figure}[t]
    \centering
    \includegraphics[width=0.9\linewidth]{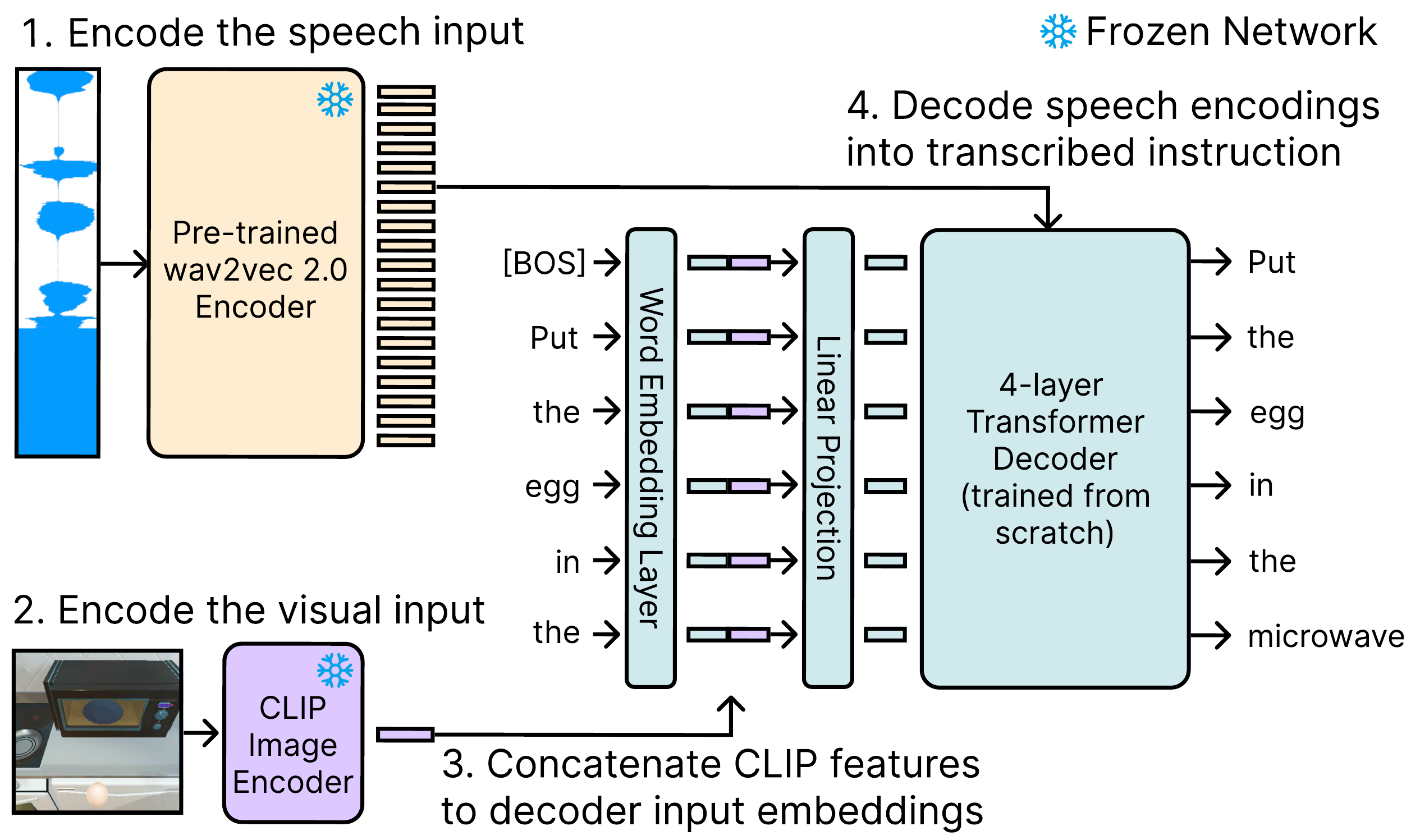}
    \caption{Our multimodal ASR model architecture. For the unimodal ASR model, the visual features are not concatenated to the word embeddings and there is no linear projection layer.}
    \label{fig:model_diagram}
\end{figure}

\subsection{Modeling}

We train unimodal and multimodal ASR models and evaluate their ability to transcribe noised speech.
The unimodal ASR model consists of a speech encoder and a language decoder. 
The speech encoder is a frozen wav2vec 2.0~\cite{baevski2020wav2vec}, pre-trained on Librispeech~\cite{panayotov2015librispeech}, which encodes the spoken instruction $S(\instruction{})$ into a sequence of speech encodings. 
The decoder is a 4-layer Transformer trained from scratch, which jointly attends over the speech encodings and decoded words.

To decouple the role of the visual modality from model architecture in ASR performance, our multimodal ASR model has a near-identical architecture (Figure~\ref{fig:model_diagram}) to the unimodal ASR model. 
The visual context $v$ is encoded using a frozen CLIP-ViT image encoder~\cite{clip} into a feature vector. 
At every timestep of generation, the CLIP feature vector is concatenated to the decoder's input word embedding, linearly projected back to the original embedding dimension, and passed through the Transformer for generating the next word.

\begin{table}[t]
    \footnotesize 
  \setlength{\tabcolsep}{3pt}
  \caption{WER (\%) and RR (\%) for ASR models trained on the American English TTS speaker across masking policies.}
  \centering
  \begin{tabular}{ llrrrrrr }
    \textbf{Test Set} & \textbf{ASR Model} & \multicolumn{1}{c}{\textbf{No Mask}} & \multicolumn{3}{c}{\textbf{Only Nouns}}&\multicolumn{2}{c}{\textbf{All Words}}\\
    & & & 20\% & 40\% & 100\% & 20\% & 40\% \\
    \toprule
    \multicolumn{8}{c}{Word Error Rate $\downarrow$}\\
    \midrule
    Seen & Unimodal& $12.6$ & $20.0$ & $26.4$ & $34.0$ & $35.2$ & $49.9$\\
    Seen & Multimodal& $\pmb{11.9}$ & $\pmb{19.9}$ & $\pmb{24.5}$ & $\pmb{30.4}$ & $\pmb{29.3}$ & $\pmb{46.4}$\\
    \midrule
    Unseen & Unimodal& $13.8$	& $\pmb{21.4}$ & $27.6$	& $34.7$	& $36.3$ & $51.7$\\
    Unseen & Multimodal& $\pmb{12.6}$ & $21.5$ & $\pmb{26.2}$ & $\pmb{33.1}$ & $\pmb{32.3}$ & $\pmb{50.2}$\\
    \\
    \multicolumn{8}{c}{Recovery Rate $\uparrow$}\\
    \midrule
    Seen & Unimodal&  -- & $61.1$ & $56.1$ & $48.0$ & $51.4$ & $38.8$\\
    Seen & Multimodal& -- & $\pmb{64.3}$ & $\pmb{60.5}$ & $\pmb{56.8}$ & $\pmb{57.7}$ & $\pmb{45.9}$\\
    \midrule
    Unseen & Unimodal& -- & $60.6$ & $55.9$ & $48.4$ & $52.0$ & $37.5$\\
    Unseen & Multimodal& -- & $\pmb{62.2}$ & $\pmb{57.5}$ & $\pmb{52.0}$ & $\pmb{53.4}$ & $\pmb{40.4}$\\
    \bottomrule
  \end{tabular}
  \label{tab:american}
\end{table}

\section{Experiments}
We train unimodal and multimodal ASR models on our spoken instructions dataset (Section~\ref{subsec:exp-details}) across several audio masking policies (Section~\ref{subsec:masking-policies}). 
We evaluate the added benefit of the visual modality when the audio signal is noised (Section~\ref{subsec:metrics}).

\subsection{Training and Inference Implementation Details}
\label{subsec:exp-details}
Across all versions of our spoken instructions dataset, we train unimodal and multimodal ASR models with cross-entropy loss. 
Models are trained for 50 epochs using the Adam optimizer with a $10^{-4}$ learning rate.
During inference, transcriptions are generated using beam search with a beam width of 5.

\subsection{Audio Masking Policies}
\label{subsec:masking-policies}
To evaluate the benefit of the visual modality with respect to the degree of acoustic noise, we conduct experiments across six noise policies for word masking.
Three masking policies consist of masking only nouns, since these words are the most visually salient. 
We identify the 100 most frequent nouns in our training set using the Natural Language Toolkit~\cite{nltk} and create three versions of our dataset where 20\%, 40\% and 100\% of these nouns in each spoken instruction $S(\instruction{})$ are masked.
Following~\cite{srinivasan2020multimodal}, we use two noise policies that mask words at random, where 20\% and 40\% of words in each $S(\instruction{})$ are masked.
Finally, a trivial masking policy is to use the original, un-noised audio.
We apply each policy at both training and inference time.

\subsection{Evaluation Metrics}
\label{subsec:metrics}

We evaluate ASR models on their Word Error Rate (WER), as well as their Recovery Rate (RR), which measures the percentage of correctly transcribed noised words in the dataset~\cite{srinivasan2020looking}. 
Further, to quantify the added benefit from the visual modality, we introduce two new metrics.

\textbf{\werr[1] (\werr)}: The relative change in Word Error Rate for the \textit{M}ultimodal model over \textit{U}nimodal.
\begin{align*}
    \text{\werr} = \frac{WER_M - WER_U}{WER_U} \times 100\%.
\end{align*}

\textbf{\rri[1] (\rri)}: The relative change in Recovery Rate for the \textit{M}ultimodal model over \textit{U}nimodal.
\begin{align*}
    \text{\rri} = \frac{RR_M - RR_U}{RR_U} \times 100\%.
\end{align*}

\section{Results and Discussion}

\begin{figure}
    \centering
    \includegraphics[width=\linewidth]{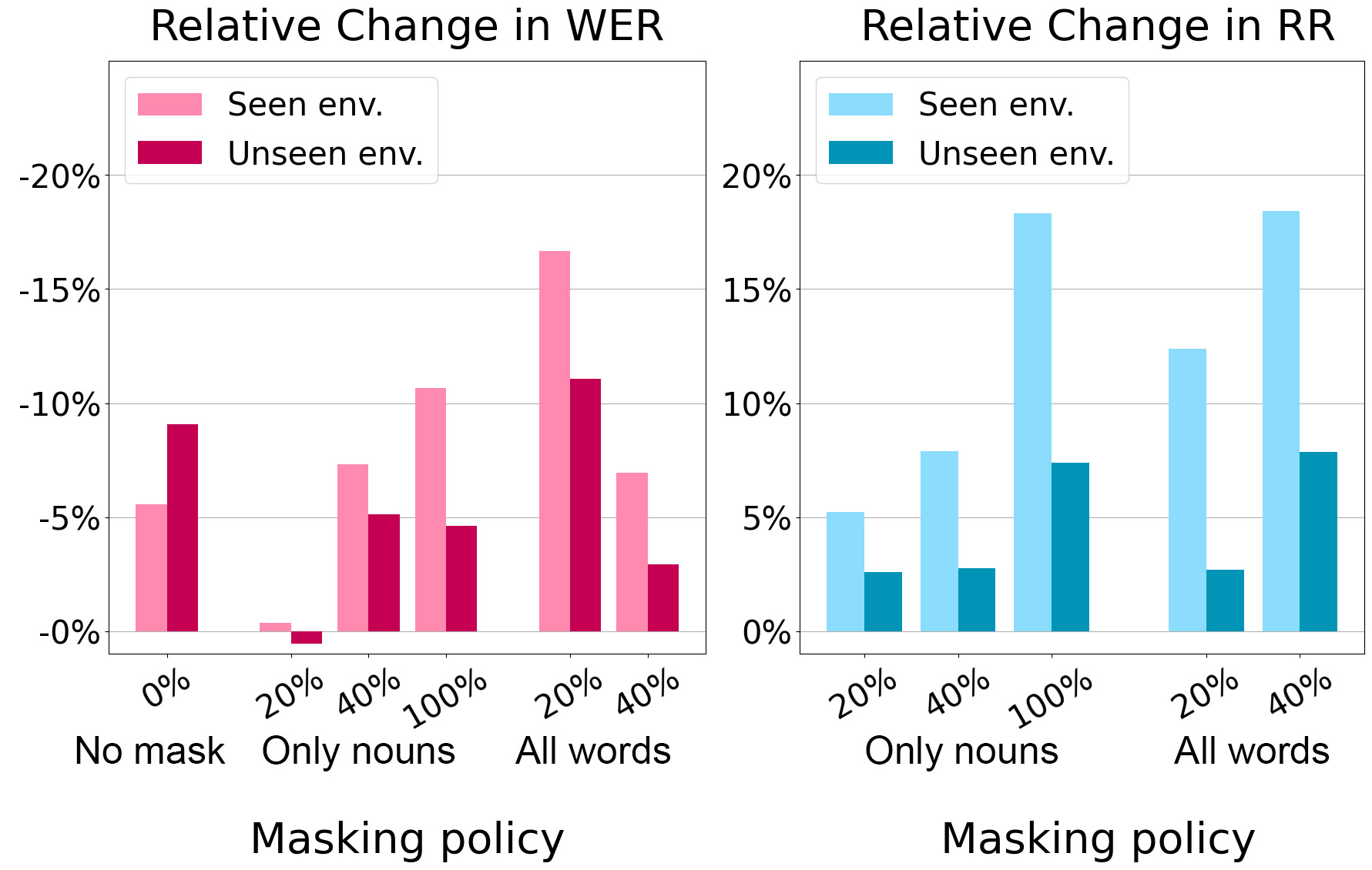}
    \caption{\werr[1] (negative is better) and \rri[1] (positive is better) between the unimodal and multimodal ASR models for the American English TTS speaker across masking policies between seen and unseen environments.}
    \label{fig:american}
\end{figure}

\subsection{Multimodal ASR Improves Masked Word Recovery in both Seen and Unseen Environments}

We test whether incorporating visual information improves ASR for spoken instructions $S_A(\instruction{})$ using a single, American English TTS model in \textit{seen} versus \textit{unseen} visual environments. 
While both unimodal and multimodal ASR models suffer from higher WER and lower RR as the level of audio masking increases, multimodal ASR models consistently outperform their unimodal counterparts across all masking policies (Table~\ref{tab:american}).

We further observe that the utility of multimodal ASR is directly proportional to the level of audio degradation, evidenced by increases in \werr and \rri as the proportion of masked words increases (Figure~\ref{fig:american}). 
This trend suggests that visual observations are more advantageous as speech signals become more degraded, and multimodal ASR may be more beneficial to agents when performing tasks in noisier environments. 
The multimodal models' WER and RR improvements generalize to unseen environments, though the improvements are less pronounced than in seen environments.
These results demonstrate the viability of this approach for embodied learning, where training and evaluation environments are often different.

\subsection{Multimodal ASR Generalizes Better to New Speakers}

\begin{figure}
    \centering
    \includegraphics[width=\linewidth]{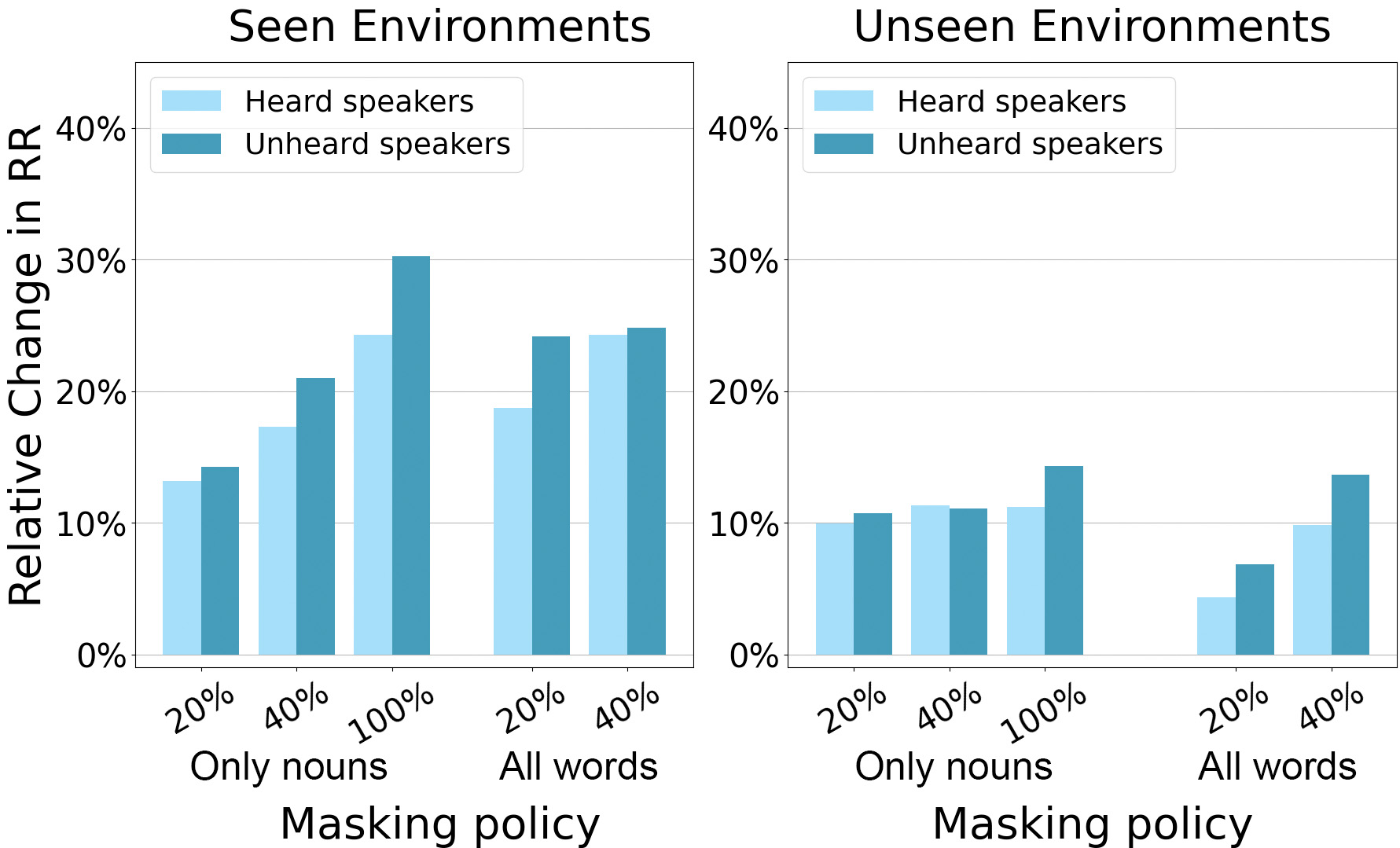}
    \caption{\rri between the unimodal and multimodal models across masking policies in both seen and unseen environments, evaluated on heard and unheard Indic English speakers.}
    \label{fig:indic_rr}
\end{figure}
\begin{table}[t]
    \footnotesize
  \setlength{\tabcolsep}{3.5pt}
  \caption{\rri on the subset of words corresponding to nouns and non-nouns from the random all-words masking policies.}
  \centering
  \begin{tabular}{ llrrrr }
    \textbf{Speaker(s)}& \textbf{POS} & \multicolumn{2}{c}{\textbf{20\% Masking}}&\multicolumn{2}{c}{\textbf{40\% Masking}}\\
    \multicolumn{2}{l}{} &\multicolumn{1}{c}{\textit{Seen}}&\multicolumn{1}{c}{\textit{Unseen}}&\multicolumn{1}{c}{\textit{Seen}}&\multicolumn{1}{c}{\textit{Unseen}}\\
    \toprule
    American & Noun & $\pmb{+12.7\%}$ & $\pmb{+03.0\%}$ & $\pmb{+31.0\%}$ & $\pmb{+14.6\%}$\\
    American & Other & $+04.4\%$ & $-02.2\%$ & $-00.1\%$ & $-01.5\%$\\
    \midrule
    Indic (Heard) & Noun & $\pmb{+20.1\%}$ & $\pmb{+05.5\%}$ & $\pmb{+40.1\%}$ & $\pmb{+18.8\%}$\\
    Indic (Heard) & Other & $-04.3\%$ & $-11.9\%$ & $+02.5\%$ & $-01.8\%$\\
    \midrule
    Indic (Unheard) & Noun & $\pmb{+25.7\%}$ & $\pmb{+07.9\%}$ & $\pmb{+45.5\%}$ & $\pmb{+24.7\%}$\\
    Indic (Unheard) & Other & $-00.7\%$ & $-08.1\%$ & $-01.6\%$ & $-00.2\%$\\
    \bottomrule
  \end{tabular}
  \label{tab:pos_rr}
\end{table}

Next, we test whether incorporating multimodal visual information improves ASR for spoken instructions of multiple speakers when those speakers are heard $S_I^{\text{heard}}(\instruction{})$ or unheard $S_I^{\text{unheard}}(\instruction{})$ during training.
Multimodal ASR is more helpful when evaluated on 5 unheard Indic English speakers, compared to the 10 Indic English speakers present in the training data (Figure~\ref{fig:indic_rr}). 
These relationships indicate that visual signals additionally regularize model training to adapt to audio from unheard speakers. 
Importantly, this relationship is observed in the joint case of \textit{unseen environments and unheard speakers}, as will be the case for newly deployed embodied agents in the world.

\subsection{Multimodal ASR is Helpful For Visually Salient Words}
We investigate whether multimodal ASR is helpful for the right reasons by evaluating whether it recovers masked words that are more visually observable. 
For the masking policies that mask words at random, we evaluate \rri on the subsets of the masked words corresponding to nouns and non-nouns. 
In Table~\ref{tab:pos_rr}, we observe that masked nouns are much more likely to be recovered than other masked words by multimodal ASR. 
\rri between nouns and non-nouns are most different in the 40\% random all-words masking policy for seen environments, suggesting that multimodal ASR is most helpful when audio is heavily perturbed but visual observations are still familiar.
The inverse is true with 20\% random all-words masking in unseen environments, which has the lowest \rri. 
These trends reveal that multimodal models are effective when masked words have strong visual salience, particularly in familiar environments.

\begin{figure}
    \centering
    \includegraphics[width=0.85\linewidth]{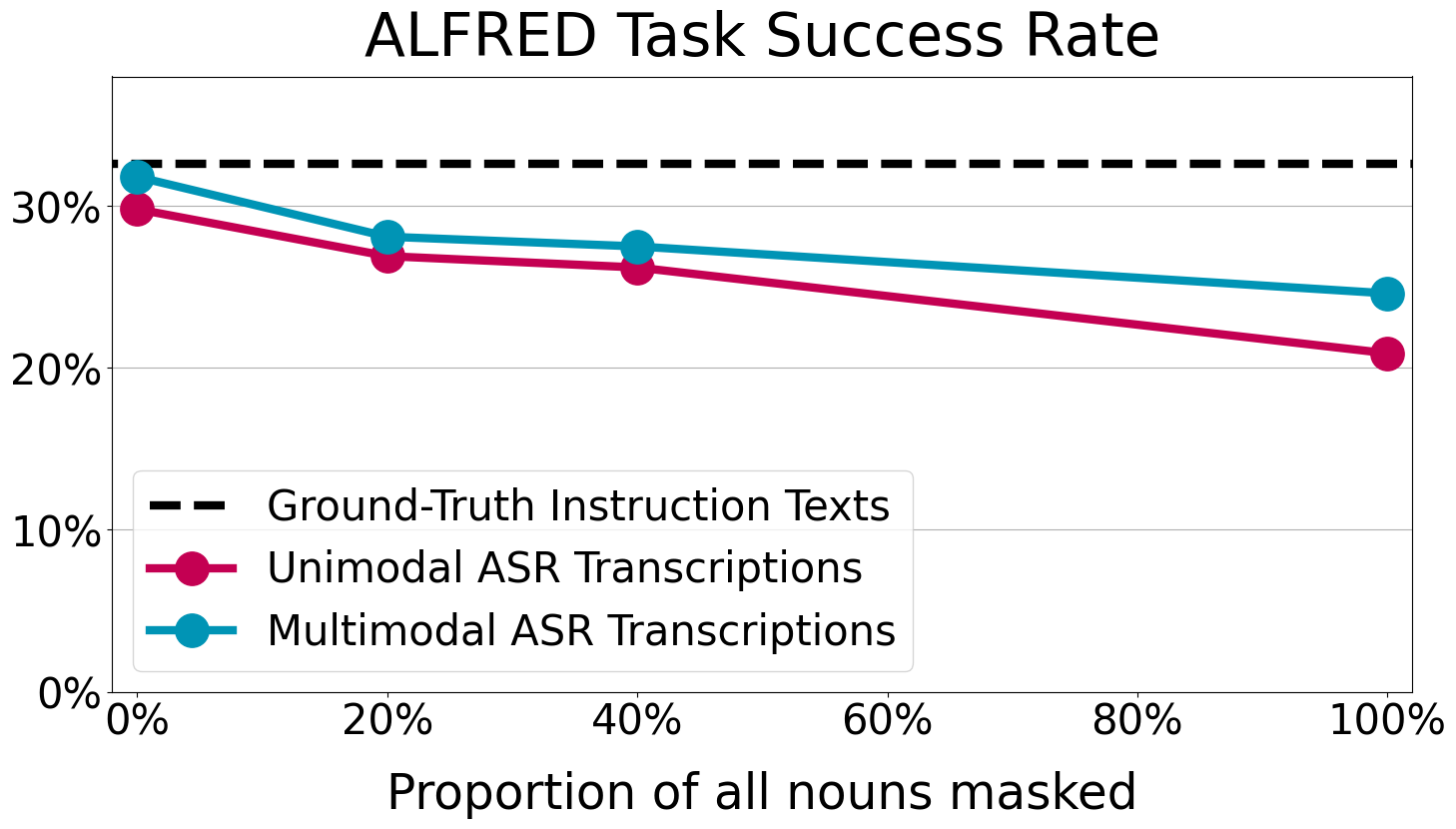}
    \caption{Text-trained ALFRED agents that observe instructions transcribed by multimodal ASR achieve higher downstream task success than when using unimodal ASR.}
    \label{fig:success_rate}
\end{figure}

\subsection{Multimodal ASR Helps Agents Complete Tasks Better}

We now investigate the impact of ASR transcription quality on an embodied agent's ability to complete tasks and whether leveraging visual observations for ASR mitigates the agent's performance degradation in the presence of noisy speech. 
We perform our analysis on the Episodic Transformer (E.T.)~\cite{pashevich2021episodic}, an off-the-shelf ALFRED agent that is trained to receive a goal text $\mathcal{G}$ and $K$ sub-goal instruction texts $\instruction{1...K}$.

We begin by extracting transcripts for the spoken instruction $S(\instruction{i})$ corresponding to every sub-goal instruction $\instruction{i}$ for each episode. 
Transcripts are extracted for the American English speaker from unimodal and multimodal ASR models, across four different masking policies (no masking and 20\%, 40\%, and 100\% noun masking). 
The ground-truth goal text $\mathcal{G}$ and the transcribed sub-goal instructions are passed to E.T., and the agent attempts to complete the task using the transcribed instructions. 
We evaluate the agent's Task Success Rate on the ALFRED validation set, in the seen environments.

We observe that transcribed instructions from both unimodal and multimodal ASR models lead to lower model performance than the ground-truth instructions (Figure~\ref{fig:success_rate}). 
However, the multimodal ASR models' transcriptions lead to lower performance degradation. 
We further observe that as the level of audio masking increases, the performance gap between the unimodal and multimodal ASR increases. 
These findings demonstrate that not only does multimodal ASR lead to more accurate transcription of spoken instructions, it also results in better downstream task completion for the embodied agent.

\section{Conclusions}
In this work, we address the challenge of embodied task
completion by following spoken instructions. 
We demonstrate that embodied agents can use their visual observations to improve ASR when transcribing spoken instructions.
ASR models using visual observations achieve better WER and RR across home environments, speakers of varied demographics, and levels of audio degradation. 
These models also show higher improvement when spoken instructions become noisier and when transcribing speech from unheard speakers, improving the robustness of ASR. 
Finally, we demonstrate that multimodal ASR can improve a pre-trained embodied agent's ability to complete tasks successfully from spoken instructions. 
These findings motivate the use of visual observations in the implementation of ASR for language-guided embodied agents.

Our work presents a proof-of-concept for spoken instruction following by creating a synthetic speech dataset and systematically masking words in the audio. 
Future work should investigate how our findings generalize to real human speakers and more realistic audio degradation.
Further, while we use only a single visual observation, future work should explore utilizing 3D scene representations built by embodied agents~\cite{film,hlsm}, as well as other strategies to actively search the visual environment, to resolve ambiguities in ASR.


\bibliographystyle{Style/IEEEtran}
\bibliography{main}

\end{document}